\documentclass[letterpaper, 10 pt, conference]{ieeeconf}  

\IEEEoverridecommandlockouts                              
                                                          
\usepackage{svg}
\usepackage{cite}
\usepackage{amsmath,amssymb,amsfonts}
\usepackage{algorithm}
\usepackage{algorithmic}
\usepackage{graphicx}
\usepackage{textcomp}
\usepackage{xcolor}
\usepackage{multirow}
\def\BibTeX{{\rm B\kern-.05em{\sc i\kern-.025em b}\kern-.08em
    T\kern-.1667em\lower.7ex\hbox{E}\kern-.125emX}}
\begin{document}

\title{\LARGE \bf
Evaluating the Benefit of Using Multiple Low-Cost Forward-Looking Sonar Beams for Collision Avoidance in Small AUVs}

\author{Christopher Morency and Daniel J. Stilwell%
\thanks{The authors are with the Bradley Department of Electrical and Computer Engineering, Virginia Polytechnic Institute and State University, Blacksburg, VA 24060, USA
        {\tt\small \{cmorency, stilwell\}@vt.edu}}%
}

\maketitle

\begin{abstract}
    We seek to rigorously evaluate the benefit of using a few beams rather than a single beam for a low-cost obstacle avoidance sonar for small AUVs. For a small low-cost AUV, the complexity, cost, and volume required for a multi-beam forward looking sonar are prohibitive. In contrast, a single-beam system is relatively easy to integrate into a small AUV, but does not provide the performance of a multi-beam solution. To better understand this trade-off, we seek to rigorously quantify the improvement with respect to obstacle avoidance performance of adding just a few beams to a single-beam forward looking sonar relative to the performance of the single-beam system. Our work fundamentally supports the goal of using small low-cost AUV systems in cluttered and unstructured environments. Specifically, we investigate the benefit of incorporating a port and starboard beam to a single-beam sonar system for collision avoidance. A methodology for collision avoidance is developed to obtain a fair comparison between a single-beam and multi-beam system, explicitly incorporating the geometry of the beam patterns from forward-looking sonars with large beam angles, and simulated using a high-fidelity representation of acoustic signal propagation.

\end{abstract}

\section{Introduction}

We address the problem of collision avoidance for an autonomous underwater vehicle (AUV) using noisy sensor data in an unknown environment using an array of inexpensive single-beam sonars. Forward-looking sonar systems for smaller AUVs usually consist of either a single-beam, such as in \cite{Calado}, or many beams, such as in \cite{Petillot}. The performance of the former is necessarily limited, and the cost and power required for the latter may be prohibitive for small AUVs. We rigorously evaluate what benefit, if any, arises from choosing a middle-ground solution that consists of a few sonar beams. Through a rigorous fundamental evaluation of the benefit of additional beams, we seek to aid in the development of a robust collision avoidance system feasible for small AUVs. 

In this work, we specifically address the obstacle avoidance problem by proposing a method for detecting obstacles and selecting avoidance maneuvers that provides a fair comparison for a forward-looking sonar system with only a few beams. We explicitly compare obstacle avoidance performance in the horizontal plane for the case of adding two additional forward-looking beams to a single-beam sonar through simulation using a high-fidelity environmental model from \cite{Morency}. The performance of the single-beam and multi-beam sonars are evaluated in environments with various object densities and sizes, at various depths and heights above the seafloor, and at varying levels of uncertainty in the dynamics of the AUV. Simulations of obstacle avoidance are in the horizontal plane in order to simplify, yet make rigorous, the comparison between the sonar systems. We present a method for obstacle mapping which explicitly incorporates sensor geometry and a reactive obstacle avoidance method using Bayesian expected loss, providing the optimal decision function for obstacle avoidance given the costs of collision and deviation from the original path.

A sonar with a low number of beams serves as a middle ground between single-beam sonars, such as the mechanically steered Imagenex 881L Profiling Sonar \cite{Heidarsson_2011} or a stationary sonar \cite{Hutin_2005}, and forward-looking imaging sonars such as DIDSON \cite{Belcher_2002} or the Blueview P450-15E \cite{Horner_2009}, which use many fixed beams to build an extensive map of the surroundings. Several approaches to mapping and collision avoidance such as in \cite{Franchi} and \cite{Teo} exist for multi-beam imaging sonars. These solutions are not practical for a system with a few beams due to the large uncertainty in object location and low resolution of sonar images.

Underwater detection and obstacle avoidance methods for AUVs is a challenging task since the system needs to be robust to the high levels of uncertainty present underwater. Several approaches to the mapping and detection problem have been presented in the literature, many of which are developed using ultrasonic sensors for indoor mobile robots such as in \cite{Hu}, \cite{Boren1990}, \cite{Fox}. These approaches have limited utility to our application due to the dynamics of an AUV since most AUVs must maintain a minimum forward velocity for depth control and have high levels of uncertainty in the dynamics. In contrast to many of the methods using ultrasonic sensors in the literature, we rigorously incorporate a physics-based sonar model to try to address some of these limitations \cite{Morency}.

A popular approach to collision avoidance in the literature is the artificial potential field \cite{Khatib}, for which obstacles are associated with repulsive forces. The forces are summed and the resultant force determines the resulting action of the AUV. The limitation of potential fields is that they require complete knowledge of the obstacles in the environment. Modifications to the potential field have been proposed by Borenstein and Koren in \cite{Boren} which can react to unexpected obstacles. However, potential fields can result in the robot becoming trapped at local minimas and oscillations, and are not practical for robots with dynamic constraints, such as AUVs. Vector field histograms \cite{Boren1990} reportedly solve some of these issues, however, most artificial potential field methods remain difficult to implement on vehicles with a restrictive turn radius \cite{Hu}. 

Our approach to the mapping problem uses a variation of an occupancy grid \cite{Elfes}. Occupancy grids are probabilistic representations of the environment for which the environment is partitioned into cells. For each measurement, cells are updated by a recursive Bayesian update. Several variations to the standard occupancy grid \cite{Ganesan}, \cite{Fulgenzi} have been proposed to improve some of the limitations of the method. Ganesan et al. \cite{Ganesan} approach the mapping problem using a local $n \times m$ occupancy grid and a motion model to propagate the grid. Using a local map improves obstacle localization by eliminating the AUV's positional error growth. The motion model is able to incorporate motion uncertainty of the obstacles relative to the AUV. In contrast, our approach uses a local occupancy grid with cells specifically constructed to match the volume ensonified by the sonar, leading to reduced computational complexity and a more accurate representation of sonar returns. Fulgenzi et al. \cite{Fulgenzi} present a method using velocity obstacles based on the Bayesian Occupancy Filter proposed by Coué et al. in \cite{Coue}. Due to the large uncertainty and beam widths of our system, using the sonar for characterizing the velocity of an object is impractical. 

A robust collision avoidance approach must be able to minimize the effect of inaccurate sensor data which can reduce the performance of a collision avoidance system. Jansson and Gustafsson \cite{Jansson} present an approach to collision avoidance using Bayes' risk for a collision mitigation system on a ground vehicle. Hu and Brady \cite{Hu} approach the collision avoidance problem from decision theory, obtaining the optimal decision rule by minimizing the Bayesian expected loss to select a minimum cost action for an industrial robot. A ground vehicle and industrial robot have the advantage of being able to stop, whereas an AUV does not. To overcome the challenges caused by dynamic constraints of an AUV, our approach utilizes Bayesian expected loss by incorporating the uncertainty in vehicle dynamics and incorporating the probability that a collision is unavoidable if no action is taken. 

The remainder of the paper is organized as follows. The detection model for mapping the environment relative to the AUV is presented in Section \ref{DM}. Our approach to reactive collision avoidance is outlined in Section \ref{CA}. Discussion of the simulator and results from Monte Carlo simulations are presented in Section \ref{SIM}.

\section{Detection Model}\label{DM}

A detection model is constructed to identify potential obstacles in the field of view of the AUV. The model maps measurements from the sonar to the probability of an obstacle at a discretized set of distances ahead of the AUV. 

\subsection{Polar Map}

\begin{figure}[htbp]
\centering 
\includegraphics[width=3.4in]{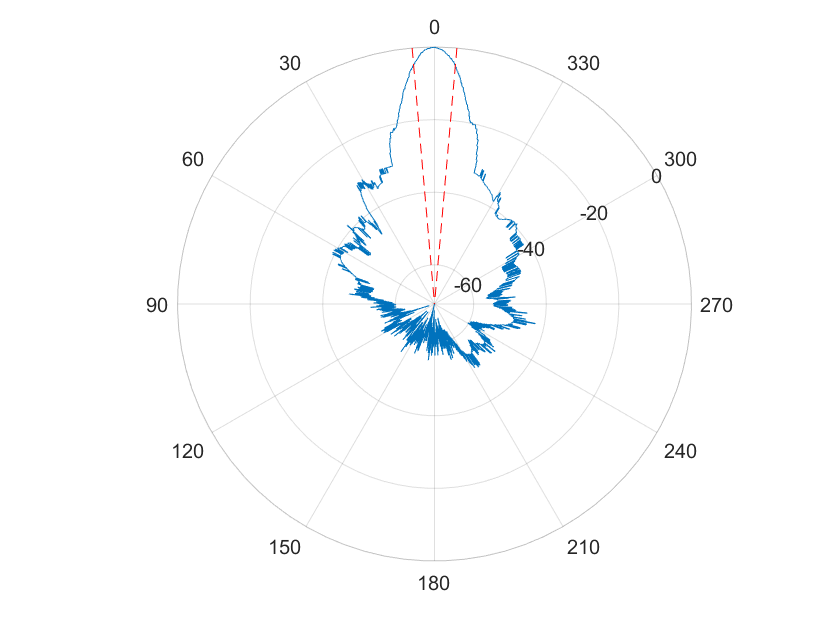}
\caption{Beam pattern loss in dB (solid blue line) and cutoff angles (dashed red line) for a single-beam forward-looking sonar}
\label{fig0}
\end{figure}

A polar map is constructed using the returns of a single-beam forward-looking sonar for which sound energy is returned to the sonar from reflections at a discrete set of times or equivalently, discrete distances. In this work, we consider a map in $\mathbb R^2$ for simplicity. Extending the framework into $\mathbb R^3$ would require additional considerations for the continuous nature of the sea floor and surface. The map is similar to the occupancy grid proposed in \cite{Elfes}, however, we explicitly adapt the cells to our sensor instead of using an $m\times n$ rectangular grid of cells. The beam pattern loss for a single-beam sonar is shown in Figure \ref{fig0}. Cut-off angles are selected such that an object between the cut-off angles is detectable 50 meters ahead of the sonar and are shown as red dotted lines at -5 and 5 degrees in Figure \ref{fig0}.

\begin{figure}[htbp]
\centering 
\includegraphics[width=2in]{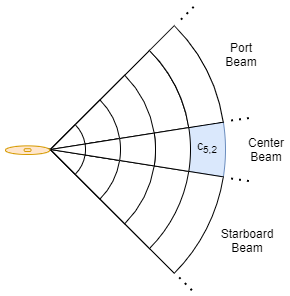}
\caption{Discretized local map using a forward-looking sonar with three beams}
\label{fig1}
\end{figure}

To characterize the returns from each beam at each of the discretized distances, a map $\mathcal{M} \in \mathbb{R}^2$ comprised of $n \times m$ cells $c_{i,j}$ is constructed such that each cell represents the probability that an obstacle is present within the bounds of the cell. The cells are constructed such that the area of each cell is the 5 dB area for a single return from the sonar for a 5 dB beam pattern loss, such that

\begin{equation*}
    c_{i,j} = \left  \{x \in \mathbb R^2 : \frac{i-1}{l_c} < x <  \frac{i}{l_c}, x \in \theta_j, \theta_{j+1} \right\}
\end{equation*}
where $l_c$ is the length of the cell in meters defined by the return from the sonar. The map is in the vehicle frame, and each cell $c_{i,j}$ comprises the area $(\frac{i-1}{l_c}, \frac{i}{l_c}]$ meters away from the vehicle and within $(\theta_j, \theta_{j+1}]$ degrees from the centerline. Figure \ref{fig1} shows an example of the discretized map using three beams where $c_{5, 2}$ corresponds to the shaded cell.

The probability of an obstacle in each cell is initially assumed to be uniform since the system has no information about the environment prior to the first sensor return. Each cell is assumed to be independent of every other cell so that the computations are tractable.

\subsection{Measurement Model}

The measurement model is used to update the map when the sensor measurements are received. The sensor receives a set of measurements $z^t \in \mathcal{Z}$, where $t$ denotes the time at which the measurement is received. The measurements are stochastic in nature, and the probability of a measurement given an obstacle, $p(z^t|\text{obstacle} \in c_{i,j}^t)$, can be computed using our measurement model \cite{Morency}. 

\subsection{Motion Model}

The motion model uses the relative rotational and translational velocity of the AUV with respect to the environment to update the obstacle map. The rotational and translational motion are decoupled to simplify computations for real-time performance. Each cell $c_{i, j} \in \mathcal M$ uses the same distribution for motion of the AUV. 

The relative translational and rotational velocity are denoted by $\tilde{v}_x$ and $\tilde{v}_r$, respectively. To propagate the map, the probability of an obstacle in each cell given the elapsed time $\tau$ between sensor measurements is computed. The probability of and obstacle in cell $c_{i,j}$ at time $t+\tau$ is

\begin{equation}\label{relative velocity eq}
\begin{split}
            &p(c_{i,j}^{t+\tau} = 1) \\ = & \bigcup_{m = 1}^{M}\bigcup_{n = 1}^{N} \bigg[p(c_{m,n}^{t} = 1) \int_{\tilde{v}\in \mathcal V} \frac{A(c_{i, j},c_{m, n}|\tilde{v}', \tau)}{A(c_{i,j}) } p(\tilde{v})d\tilde{v}'\bigg] 
\end{split}
\end{equation}
where $A\left(c_{i, j},c_{m, n}|v, \tau\right)$ is the overlapping area of cells $c_{i,j}$ and $c_{m, n}$ at time $t+ \tau$ given velocity $v$, $A(c_{m,n})$ is the total area of cell $c_{m, n}$, and $\mathcal V$ is the velocity distribution. 

\begin{figure}[htbp]
\centering 
\includegraphics[width=3.4in]{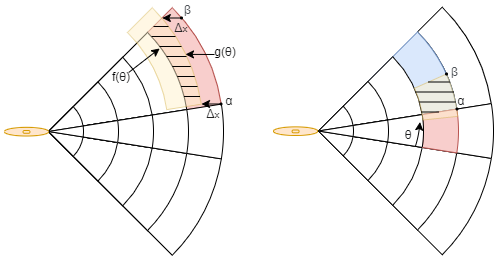}
\caption{Translational propagation (left) and rotational propagation (right)}
\label{fig2}
\end{figure}

Since each cell is independent and a constant velocity distribution is used, the unions in \eqref{relative velocity eq} can be written as sums, thus, 
\begin{equation}\label{rel vel 2}
\begin{split}
        &p(c_{i,j}^{t+\tau} = 1) \\ = &\sum_{m = 1}^{M}\sum_{n = 1}^{N} \bigg[p(c_{m,n}^{t} = 1) \int_{\tilde{v}\in \mathcal V} \frac{A(c_{i, j},c_{m, n}|\tilde{v}', \tau)}{A(c_{i,j}) } p(\tilde{v})d\tilde{v}'\bigg]
\end{split}
\end{equation}
For real-time computation, the translational and rotational motion are decoupled and \eqref{rel vel 2} is first computed with $\mathcal V$ as the distribution for translational motion. For a given translational velocity, $v_x$, the overlapping area of two cells is 

\begin{equation}
    A\left(c_{i, j},c_{m, n}|v_x, \tau \right) = \int_\alpha^\beta \int_{f(\theta, \tau)}^{g(\theta, \tau)} r drd\theta
\end{equation}
where $f(\theta, \tau)$ and $g(\theta, \tau)$ are the bounds of the overlapping cell given $\theta$ and $\tau$. The angles $\alpha$ and $\beta$ are the minimum and maximum angles, respectively, for which a point in cell $c_{m,n}$ at that angle overlaps with cell $c_{i,j}$. Figure \ref{fig2} shows an illustration on the left for computing the area of a cell which overlaps with itself after a displacement given by $\Delta x = v_x \tau$. 

The overlapping area, $A(c_{i, j},c_{m, n}|\omega, \tau)$ given rotational velocity $\omega$ over time $\tau$ is

\begin{equation*}
    A\left(c_{i, j},c_{m, n}|\omega, \tau\right) = \begin{cases} \frac{\beta - \alpha}{\gamma}  & i = m \\ 0 & i \neq m\end{cases}
\end{equation*}
where $\alpha$ and $\beta$ are the minimum and maximum overlap angles, respectively, and $\gamma$ is the total width of the shifted cell $c_{m, n}$. Figure \ref{fig2} shows an example computation on the right for computing the overlap of an adjacent cell, given an angular rotation $\theta$. 


The translational and rotational velocity distributions can be chosen using navigation data from the vehicle, when available. If navigation data is unknown, a distribution should be selected with a large variance.

\subsection{Bayesian update}

A Bayesian update is performed to combine new sets of measurements, $z^t$, received by the sonar. For each cell $c_{i,j} \in \mathcal{M}$, 

\begin{equation*}
        p\left(c_{i,j}^{t} = 1| z^t\right) = \frac{p(z^t| c_{i,j}^{t} = 1)p(c_{i,j}^{t} = 1)}{\sum_{k = 0}^1 p(z^t|c_{i,j}^{t} = k)p(c_{i,j}^{t} = k)}
\end{equation*}
where $p(z^t|c_{i,j}^{t} = 1)$ is from the sensor model and $p(c_{i,j}^{t} = 1)$ is the prior which is given by the propagated probabilities from the velocity distributions. The probability of no obstacle in a cell is $p(c_{i,j}^{t} = 0) = 1 - p(c_{i,j}^{t} = 1)$.

\subsection{Sonar Model}

A high-fidelity sonar model \cite{Morency} is used to model the acoustic propagation of the sonar. The model consists of a set of equations for the sound velocity \cite{Medwin_1975}, attenuation \cite{z1}, \cite{z2}, spread loss \cite{Jenson_2011}, beam pattern loss using the single-point-source approach \cite{Marage_2010}, backscatter \cite{Urick_1983}, sonar resolution \cite{Medwin_1998} and noise \cite{Coates_1990}. The backscatter is computed for the seafloor, sea surface, and volume. The noise sources considered are turbulence, shipping traffic, sea state and thermal noise.

\section{Collision Avoidance}\label{CA}

In this section, we present a reactive collision avoidance approach. In Section \ref{CA1}, we present a loss function and a decision rule for when the AUV should react to an obstacle given a pencil beam sonar. In Section \ref{CA2} we present a framework for the case with the additional port and starboard beams where multiple actions are available. Our approach uses the posterior expected loss which, given a map and loss functions defined for each available maneuver, gives the optimal decision function. The framework presented in Sections \ref{CA1} and \ref{CA2} is based on the standard decision-theory framework for deciding when to react, as proposed in \cite{Hu} and \cite{Jansson}. In Section \ref{CA3}, we extend this standard framework to incorporate a probabilistic trajectory. 

\subsection{Reaction Decision Rule}\label{CA1}

A collision in the time frame $[t, t+\tau]$, called our threat assessment interval, is represented by an indicator function 

\begin{equation*}
    C(X_{t:t+\tau}) = \begin{cases} 0 & \text{no collision} \\ 1 & \text{collision} \end{cases}
\end{equation*}
where $X_{t:t+\tau}$ is the trajectory of the AUV. 

When there is no collision in the window, there could be a collision which will be unavoidable past the window, due to limitations of vehicle maneuverability. To represent a collision becoming unavoidable within the threat assessment interval, an indicator function is used

\begin{equation*}
        C_U(X_{t:t+\tau}) = \begin{cases} 0 & \text{no unavoidable collision} \\ 1 & \text{unavoidable collision} \end{cases}
\end{equation*}

The probability of collision given a set of measurements is a stochastic process that can be written as a hypothesis test,

\begin{align*}
    H_0 : C_U(X_{t:t+\tau}) = 0 \\
    H_1 : C_U(X_{t:t+\tau}) = 1
\end{align*}
where $\Theta = \{H_0, H_1\}$ is our state space. 
The decision rule determines if an intervention is necessary. Posterior expected loss is used to determine the optimal decision rule. The cost of no intervention is $R(H_0| z_{0:t})$ and the cost of an intervention is $R(H_1| z_{0:t})$ where

\begin{equation*}
    R(H_i| z_{0:t}) = \sum_{j = 0}^1 L(H_i, H_j)P(H_j| z_{0:t})
\end{equation*}
The optimal decision rule given our loss function selects the minimum risk for these two actions, thus, 

\begin{equation*}
    \delta(X) = 1 \text{  if  } R(H_0| z_{0:t}) > R(H_1| z_{0:t})
\end{equation*}
where $\delta(X) = 1$ is an intervention.
The two decision rules are each binary, leading to four possible scenarios and their associated costs,

\begin{equation*}
    L(H_i, H_j) = \begin{cases}
    C_{0, 0} & \text{Correct non-intervention} \\
    C_{0,1} & \text{Collision} \\
    C_{1, 0} & \text{False alarm} \\
    C_{1,1} & \text{Correct intervention}
    \end{cases}
\end{equation*}
where $C_{H_i, H_j}$ is the cost of assuming $H_i$ given $H_j$. The costs of correct intervention and false alarm incorporate a prior probability for an intervention that results in collision due to an additional obstacle outside the field of view of the sonar. The posterior expected loss for non-intervention and intervention are $R(H_0| z_{0:t})$ and $R(H_1| z_{0:t})$, respectively, where 

\begin{equation}\label{PEL1}
    R(H_i| z_{0:t}) = C_{i, 0}P(H_0| z_{0:t}) + C_{i, 1}P(H_1| z_{0:t})
\end{equation}
The decision rule selects the action with the least posterior expected loss, thus, if $R(H_0| z_{0:t}) > R(H_1| z_{0:t})$, an intervention is performed. Substituting formula \eqref{PEL1}, 

\begin{equation*}
        R(H_0| z_{0:t}) > R(H_1| z_{0:t}) \implies \frac{P(H_1| z_{0:t})}{P(H_0| z_{0:t})} > \frac{C_{1, 0} - C_{0, 0}}{C_{0,1} - C_{1, 1}}
\end{equation*}
Using this decision rule is useful if a single forward-looking pencil beam sonar is used, however, if further knowledge of the environment is available either a priori or from an additional sensor, this method does not incorporate the additional knowledge.
\subsection{Multiple Actions}\label{CA2}

If further knowledge of the environment is known, it can be incorporated into the decision process by the collision avoidance system. Denote the action space $\mathcal{A}$, for which action $k$ is denoted by $a^k$. The action space can simply be a decision on whether to intervene, such as for the pencil beam sonar described in the previous section, or it can combine multiple possible actions, such as 

\begin{equation*}
    \mathcal{A} = \begin{cases} a^0 & \text{go straight} \\ a^1 & \text{turn left} \\ a^2 & \text{turn right} \end{cases}
\end{equation*}
The posterior expected loss given that the AUV takes action $a^k$ is

\begin{equation*}
    R(a^k| z_{0:t}) = \sum_{j = 0}^1 L(a^k, H_j)P(H_j| z_{0:t})
\end{equation*}
The decision function is modified to take action $a^*$, 

\begin{equation*}
    a^* = \underset{a^i}{\text{argmin }} R(a^i| z_{0:t}) = \underset{a^k}{\text{argmin}} \sum_{j = 0}^1 L(a^k, H_j)P(H_j| z_{0:t})
\end{equation*}
which is the optimal action to minimize the overall posterior expected loss.


The loss function determines the trade-off between the cost of a collision and the cost of a false alarm or unnecessary maneuver. The loss for action $a^k$ given $H_0$ is $C_0^k$, where $C_0^k$ is the cost for taking the trajectory from action $k$, for which an example metric could be the additional distance from the end goal. The loss for action $a^k$ given $H_1$ is $C_0^k + C_1^k$, where $C_1^k$ is the cost of collision for action $k$ and $C^k_{1} = C^j_{1}$ $\forall j, k$ since the cost of collision is the same for each action. To tune based on the cost of false alarm, the cost is $C_{0}^k - C_0^0$ $\forall k \neq 0$ where the cost for a missed detection (collision) is $C_{1}^0 + C_0^0$. 

 \begin{figure}[t]
\centering 
\includegraphics[width=3.4in]{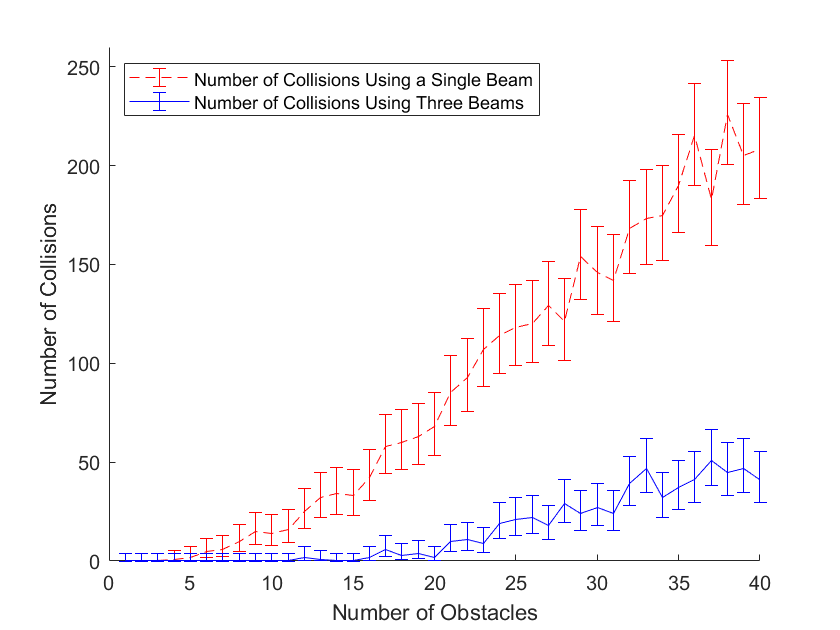}
\caption{Collisions for 1000 runs with varying numbers of obstacles with 95\% confidence intervals}
\label{fig10}
\end{figure}
\subsection{Probabilistic Trajectory}\label{CA3}

We use a stochastic approximation of the AUV dynamics such that for some desired action, the AUV trajectory will be one of the possible trajectories $X_{t:t+T}\in \mathcal X$ where $\mathcal X$ is the set of all possible trajectories. The probability $p(X_{t:t+T})$ that the AUV trajectory is $X_{t:t+T}$ is dependent on the selected action. The decision function is constructed to estimate the probability of a collision given a command over the possible trajectories. Given a command to perform action $a^k \in \mathcal A$, the trajectory incorporates a variance to account for positional uncertainty. The cells in the trajectory of the AUV given action $a^k$ over $T$ seconds is $X_{t:t+T}^k$. The probability that the trajectory passes through a cell $c_{i,j}$ within $T$ seconds given action $a^k$ is 

\begin{equation*}
    p(c_{ij} \in X_{t:t+T}|a^k)   = \int_{X^k_{t:t+T} \in \mathcal{X}} I(x \in c_{i, j})p(x|a^k)dx
\end{equation*}
where $I(x \in c_{i, j})$ is an indicator function which is 1 if the trajectory passes through cell $c_{i, j}$. 

The probability of collision given an action $a^k$ incorporates the probability that the trajectory passes through each cell and the probability of each cell containing an obstacle. Thus, 

\begin{equation*}
\begin{split}
    p&(C(X_{t:t+T}) = 1| a^k) \\ &= \sum_{i} \sum_{j} p(c_{i, j} = 1)p(X_{t:t+T} \in c_{i, j}| a^k) 
\end{split}
\end{equation*}
The probability of collision makes the assumption that if a cell $c_{i,j}$ contains an obstacle, passing through cell $c_{i, j}$ will result in a collision.

\begin{figure}[t]
\centering 
\includegraphics[width=3.4in]{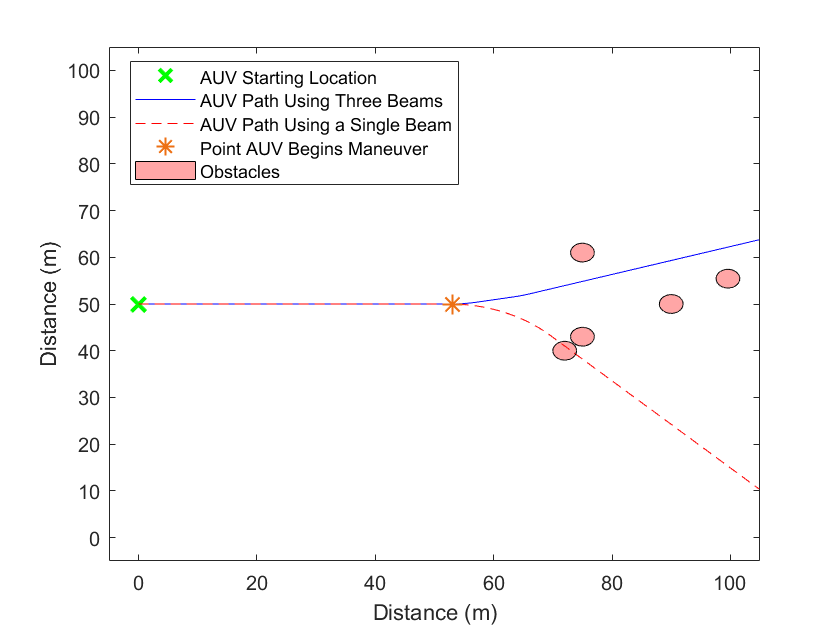}
\caption{Sample collision avoidance run}
\label{fig6}
\end{figure}

\section{Simulations}\label{SIM}

To evaluate the performance of our method using a local occupancy grid and motion model, Monte Carlo simulations are performed comparing a single-beam sonar system to a three-beam sonar system. Simulations are performed comparing the effects of dense obstacle fields, uncertainty in yaw and surge velocity, height above the seafloor, and varying the size of obstacles. The maneuvers for the AUV are in the horizontal plane, however, returns from the seafloor and sea surface are incorporated. To extend the collision avoidance framework to the vertical plane, an approach incorporating the continuous nature of the seafloor and surface would be beneficial.

The forward beam has a 5 dB beam width of 10 degrees as shown in Figure \ref{fig0}, and both the port and starboard beams have a 5 dB beam width of 20 degrees. The centers of the port and starboard beams are offset at -15 and 15 degrees, respectively. A high-fidelity environmental model from \cite{Morency} is used to simulate the returns to the sonar, which models effects from sound velocity, transmission loss, beam patterns, noise, sonar resolution, and backscatter from the seafloor, sea surface, and volume. The AUV motion is approximated with a simple unicycle model such that the velocity $\mathcal U$ is 

\begin{equation*}
    \mathcal U = \begin{bmatrix} \dot x \\ \dot y \\ \dot \phi \end{bmatrix} = \begin{bmatrix} V \cos \phi \\ V \sin \phi \\ \omega \end{bmatrix}
\end{equation*}
where $x$ and $y$ are the AUV positions in x-y coordinates, $V$ is the surge velocity, $\phi$ is the heading angle and $\omega$ is the rotational velocity. The collision avoidance methods described in Section \ref{CA} are used for both the single and multi-beam simulations. 

The simulation environment is 100 $\times$ 100 meters with obstacles randomly placed between 50 and 100 meters ahead of the AUV. The target strength of the obstacles is selected such that they are detected by the sensor 50 meters ahead of the AUV, and have a minimum target strength of -30 dB. Simulations are run in $\mathbb R^2$ but can be extended to $\mathbb R^3$. We perform Monte Carlo runs for each simulation on randomly generated maps, with each map run once using a single-beam and once using three beams. 

\begin{figure}[t]
\centering 
\includegraphics[width=3.4in]{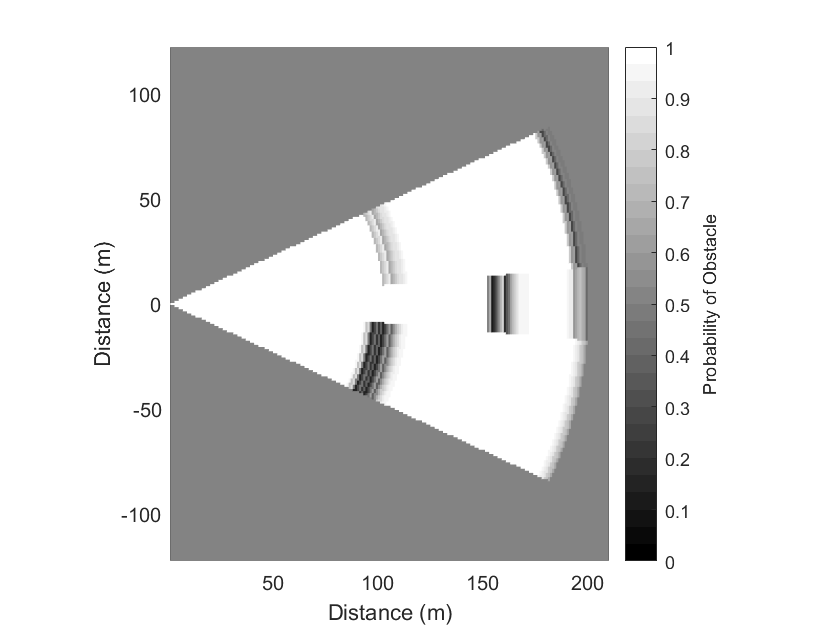}
\caption{Local map with three beams}
\label{fig8}
\end{figure}

Figure \ref{fig10} contains the results of the simulations using various object densities, run 1000 times for each density. The AUV is simulated 25 meters above the seafloor and 50 meters below the sea surface, with no uncertainty in the AUV dynamics, a surge velocity of 5 knots, and each obstacle has a radius of 2 meters. The number of collisions using a single-beam sonar is shown as a red dotted line and the number of collisions using three forward-looking beams is a solid blue line. The 95\% confidence intervals are computed using the Clopper-Pearson method. In sparse environments with few obstacles ahead of the AUV, both the single and multi-beam sonars successfully avoid the obstacles. When more than 3 obstacles are present, using the additional two beams outperforms using a single-beam. The additional two beams outperform the single-beam sonar by more than a factor of 10 when there are 10 obstacles, however, when more than 20 obstacles are added to the environment, both methods fail more frequently and the sonar with two additional beams outperforms the single-beam by a smaller factor.

\begin{figure}[t]
\centering 
\includegraphics[width=3.4in]{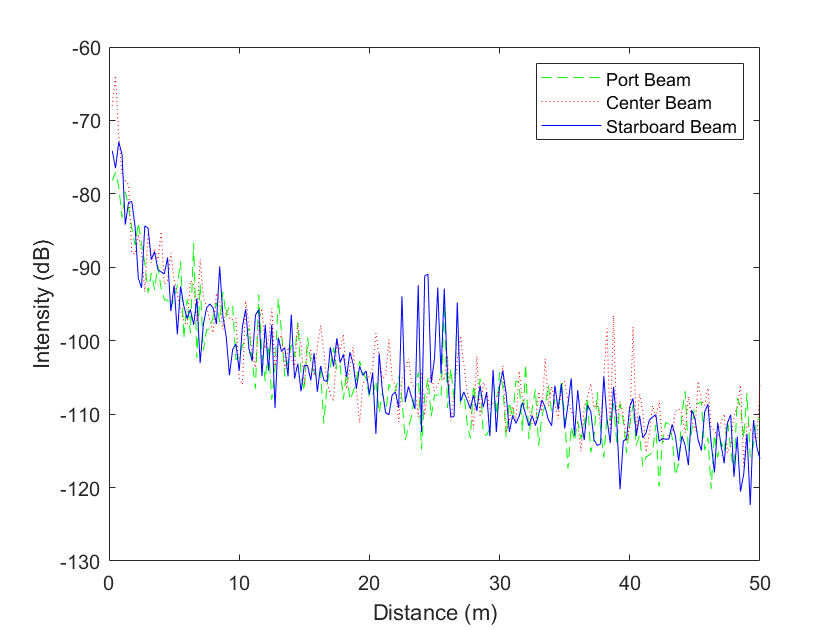}
\caption{Beam returns from sample collision avoidance run}
\label{fig9}
\end{figure}

A sample Monte Carlo run in a randomly generated map is shown in Figure \ref{fig6} where the path of the AUV taken using a single-beam is shown as a red dotted line, the path taken by the AUV using the two additional beams is shown as the blue solid line, and obstacles are denoted by red circles. The point immediately before the AUV with three beams turns left is denoted by an orange asterisk in Figure \ref{fig6}. The local map of the AUV with three beams at the orange asterisk is shown in Figure \ref{fig8}, where darker cells represent higher probability of an obstacle. The additional two beams can detect that the AUV is significantly less likely to result in collision by maneuvering to the left, as shown in Figure \ref{fig8}, allowing the AUV to avoid a potential collision, whereas the single-beam only detects the obstacle directly ahead, resulting in a collision. Figure \ref{fig9} shows the simulated returns from each beam when the AUV is at the orange asterisk in Figure \ref{fig6} where the returns from the port beam is a green dashed line, the center beam is a red dotted line, and the starboard beam is a solid blue line.

Varying the size of objects can affect the performance of the obstacle avoidance algorithm, since smaller obstacles require smaller maneuvers to avoid. Simulations are run with 10 obstacles each with a radius between 0.5 meters and 4 meters, with 1000 runs for each radius. The AUV has a surge velocity of 5 meters, 25 meters above the seafloor. The results of the simulations are shown in Table \ref{tab:tab5}. As the size of the obstacles increase, the number of collisions for both the AUV with a single-beam and the AUV with multiple beams increase. When objects have a 4 meter radius the additional beams outperform the single-beam by a factor of 3.25 compared to a factor of 9.5 when objects have a 2 meter radius.

\begin{table}[htbp]
\renewcommand{\arraystretch}{1.3}
\caption{Monte Carlo Runs of Simulation Varying Object Size}
\label{tab:tab5}
\centering
\begin{tabular}{|p{70pt}|p{50pt}|c|}
    \hline
      & \multicolumn{2}{|c|}{Number of Collisions}  \\
    \hline
    Object Radius (m) & Single-Beam & Three Beams  \\
    \hline
    \hline
    0.5 & 0 & 0 \\
    1 & 3 & 0 \\   
    2 & 19  & 2\\   
    3 & 28 & 5\\
    4 & 39 & 12\\
    \hline
\end{tabular}
\end{table}

Small AUVs may not directly measure ground relative velocity, leading to uncertainty in surge velocity. Simulations are performed with varying uncertainty in surge velocity, using 10 obstacles each with a radius of 2 meters. The AUV's estimate of its surge velocity is 5 knots. Results of the simulations show that if the true velocity of the AUV is within 5 knots of the estimated velocity, which is far beyond a reasonable velocity estimation error, there is no change in obstacle avoidance performance for both the single-beam and multi-beam case. Uncertainty in rotational velocity is also simulated with an error of 0.01 rad/s. No change in performance was found for either the single-beam or multi-beam case. The results of these simulations show that our algorithm is feasible for a low-cost navigation solution.

Our current analysis is limited to the horizontal plane, although the simulation environment fully captures the effect of the seafloor on sonar returns. Simulations are performed with the AUV at varying heights above seafloor. When the height above the seafloor is small, side beams have significantly more uncertainty at further distances, leading to higher false alarm rates. Since the center beam is small, there is less backscatter from the seafloor keeping false positives to a minimum. We perform 1000 Monte Carlo runs for various heights between 3 meters and 50 meters above the seafloor, with no uncertainty in the AUV dynamics, a surge velocity of 5 knots, and 10 obstacles with a radius of 2 meters. No change in performance at different heights above the seafloor was found with either the single-beam or multi-beam sonars.

\section{Conclusion}

Our analysis shows that adding two additional beams outperforms using a single-beam when more than 3 obstacles are present in the environment and outperforms a single-beam by more than a factor of 10 when there are 10 objects present in the environment. The proposed solution is shown to work 3 meters or higher above the seafloor and is useful for small AUVs with high uncertainty in surge velocity and yaw. Looking forward, experimentation on a real system will help validate this approach for practical use. 

\bibliographystyle{IEEEtran}
\bibliography{IEEEabrv,bibi}

\begin{thebibliography}{10}
\providecommand{\url}[1]{#1}
\csname url@rmstyle\endcsname
\providecommand{\newblock}{\relax}
\providecommand{\bibinfo}[2]{#2}
\providecommand\BIBentrySTDinterwordspacing{\spaceskip=0pt\relax}
\providecommand\BIBentryALTinterwordstretchfactor{4}
\providecommand\BIBentryALTinterwordspacing{\spaceskip=\fontdimen2\font plus
\BIBentryALTinterwordstretchfactor\fontdimen3\font minus
  \fontdimen4\font\relax}
\providecommand\BIBforeignlanguage[2]{{%
\expandafter\ifx\csname l@#1\endcsname\relax
\typeout{** WARNING: IEEEtran.bst: No hyphenation pattern has been}%
\typeout{** loaded for the language `#1'. Using the pattern for}%
\typeout{** the default language instead.}%
\else
\language=\csname l@#1\endcsname
\fi
#2}}

\bibitem{Calado}
P.~Calado, R.~Gomes, M.~B. Nogueira, J.~Cardoso, P.~Teixeira, P.~B. Sujit, and
  J.~B. Sousa, ``Obstacle avoidance using echo sounder sonar,'' in \emph{OCEANS
  2011 IEEE - Spain}, 2011, pp. 1--6.

\bibitem{Petillot}
Y.~Petillot, I.~Tena~Ruiz, and D.~Lane, ``Underwater vehicle obstacle avoidance
  and path planning using a multi-beam forward looking sonar,'' \emph{IEEE J.
  Ocean. Eng.}, vol.~26, no.~2, pp. 240--251, 2001.

\bibitem{Morency}
C.~Morency, D.~J. Stilwell, and S.~Hess, ``Development of a simulation
  environment for evaluation of a forward looking sonar system for small
  {AUVs},'' in \emph{OCEANS 2019 MTS/IEEE SEATTLE}, 2019, pp. 1--9.

\bibitem{Heidarsson_2011}
H.~K. Heidarsson and G.~S. Sukhatme, ``Obstacle detection and avoidance for an
  {Autonomous} {Surface} {Vehicle} using a profiling sonar,'' in \emph{Proc.
  IEEE Int. Conf. Robot. Autom.}, 2011, pp. 731--736.

\bibitem{Hutin_2005}
E.~Hutin, Y.~Simard, and P.~Archambault, ``Acoustic detection of a scallop bed
  from a single-beam echosounder in the {St. Lawrence},'' \emph{ICES J. Mar.
  Sci.}, vol.~62, no.~5, pp. 966--983, 2005.

\bibitem{Belcher_2002}
E.~Belcher, W.~Hanot, and J.~Burch, ``Dual-frequency identification sonar
  ({DIDSON}),'' in \emph{Proc. Int. Symp. Underwater Technology}, 2002, pp.
  187--192.

\bibitem{Horner_2009}
D.~Horner, N.~McChesney, T.~Masek, and S.~Kragelund, ``{3D} reconstruction with
  an {AUV} mounted forward looking sonar,'' in \emph{Proc. Int. Symp. on
  Unmanned Untethered Submersible Technology}, Durham, NH, 2009.

\bibitem{Franchi}
M.~Franchi, A.~Bucci, L.~Zacchini, E.~Topini, A.~Ridolfi, and B.~Allotta, ``A
  probabilistic {3D} map representation for forward-looking sonar
  reconstructions,'' in \emph{2020 IEEE/OES Auton. Underw. Veh. Symp.}, 2020,
  pp. 1--6.

\bibitem{Teo}
K.~Teo, K.~W. Ong, and H.~C. Lai, ``Obstacle detection, avoidance and anti
  collision for {MEREDITH AUV},'' in \emph{OCEANS 2009}, 2009, pp. 1--10.

\bibitem{Hu}
H.~{Hu} and M.~{Brady}, ``A {Bayesian} approach to real-time obstacle avoidance
  for a mobile robot,'' \emph{Auton. Robots}, vol.~1, no.~1, pp. 1573--7527,
  1994.

\bibitem{Boren1990}
J.~Borenstein and Y.~Koren, ``Real-time obstacle avoidance for fast mobile
  robots in cluttered environments,'' in \emph{Proc. IEEE Int. Conf. Robot.
  Autom.}, 1990, pp. 572--577 vol.1.

\bibitem{Fox}
D.~Fox, W.~Burgard, and S.~Thrun, ``The dynamic window approach to collision
  avoidance,'' \emph{IEEE Robot. Autom. Mag.}, vol.~4, no.~1, pp. 23--33, 1997.

\bibitem{Khatib}
O.~Khatib, ``Real-time obstacle avoidance for manipulators and mobile robots,''
  in \emph{Proc. IEEE Int. Conf. Robot. Autom.}, vol.~2, 1985, pp. 500--505.

\bibitem{Boren}
J.~Borenstein and Y.~Koren, ``Real-time obstacle avoidance for fast mobile
  robots,'' \emph{IEEE Trans. on Systems, Man, and Cybernetics}, vol.~19,
  no.~5, pp. 1179--1187, 1989.

\bibitem{Elfes}
A.~Elfes, ``Using occupancy grids for mobile robot perception and navigation,''
  \emph{Computer}, vol.~22, no.~6, pp. 46--57, 1989.

\bibitem{Ganesan}
V.~{Ganesan}, M.~{Chitre}, and E.~{Brekke}, ``Robust underwater obstacle
  detection and collision avoidance,'' \emph{Auton. Robots}, vol.~40, no.~7,
  pp. 1165--1185, 2016.

\bibitem{Fulgenzi}
C.~Fulgenzi, A.~Spalanzani, and C.~Laugier, ``Dynamic {Obstacle} {Avoidance} in
  uncertain environment combining {PVOs} and {Occupancy} {Grid},'' in
  \emph{Proc. IEEE Int. Conf. Robot. Autom.}, 2007, pp. 1610--1616.

\bibitem{Coue}
C.~Coué, C.~Pradalier, C.~Laugier, T.~Fraichard, and P.~Bessière, ``Bayesian
  occupancy filtering for multitarget tracking: an automotive application,''
  \emph{Int. J. Robot. Res.}, vol.~25, no.~1, pp. 19--30, 2006.

\bibitem{Jansson}
J.~{Jansson} and F.~{Gustafsson}, ``A framework and automotive application of
  collision avoidance decision making,'' \emph{Automatica}, vol.~44, no.~9, pp.
  2347--2351, 2008.

\bibitem{Medwin_1975}
H.~Medwin, ``Speed of sound in water: A simple equation for realistic
  parameters,'' \emph{J. Acoust. Soc. Am.}, vol.~58, no.~6, pp. 1318--1319,
  1975.

\bibitem{z1}
R.~E. Francois and G.~R. Garrison, ``Sound absorption based on ocean
  measurements: Part i: Pure water and magnesium sulfate contributions,''
  \emph{J. Acoust. Soc. Am.}, vol.~72, no.~3, pp. 896--907, 1982.

\bibitem{z2}
------, ``Sound absorption based on ocean measurements. part ii: Boric acid
  contribution and equation for total absorption,'' \emph{J. Acoust. Soc. Am.},
  vol.~72, no.~6, pp. 1879--1890, 1982.

\bibitem{Jenson_2011}
F.~B. Jensen, W.~A. Kuperman, M.~B. Porter, and H.~Schmidt, \emph{Computational
  Ocean Acoustics}.\hskip 1em plus 0.5em minus 0.4em\relax New York: Springer,
  2011.

\bibitem{Marage_2010}
J.~P. Marage and Y.~Mori, \emph{Sonar and Underwater Acoustics}.\hskip 1em plus
  0.5em minus 0.4em\relax London: ISTE, 2010.

\bibitem{Urick_1983}
R.~Urick, \emph{Principles of Underwater Sound}, 3rd~ed.\hskip 1em plus 0.5em
  minus 0.4em\relax New York: McGraw-Hill, 1983.

\bibitem{Medwin_1998}
H.~Medwin and C.~S. Clay, \emph{Fundamentals of Acoustical Oceanography}.\hskip
  1em plus 0.5em minus 0.4em\relax Boston, MA: Academic Press, 1998.

\bibitem{Coates_1990}
R.~F.~W. Coates, \emph{Underwater Acoustic Systems}.\hskip 1em plus 0.5em minus
  0.4em\relax New York: John Wiley \& Sons, 1989.

\end{thebibliography}

\end{document}